\documentclass{article}

\usepackage[final]{new_in_ML}

\usepackage[utf8]{inputenc} 
\usepackage[T1]{fontenc}    
\usepackage{hyperref}       
\usepackage{url}            
\usepackage{booktabs}       
\usepackage{amsfonts}       
\usepackage{nicefrac}       
\usepackage{microtype}      
\usepackage{xcolor}         
\usepackage{enumitem}

\usepackage{graphicx}
\usepackage{subcaption} 

\title{FoodFusion: A Latent Diffusion Model for Realistic Food Image Generation}

\author{
Olivia Markham\thanks{Equal Contribution}\\
Systems Design Engineering\\
University of Waterloo\\
Waterloo, ON\\
ogmarkha@uwaterloo.ca\\
\And
Yuhao Chen\thanks{Equal Contribution}\\
Systems Design Engineering\\ 
University of Waterloo\\
Waterloo, ON\\
yuhao.chen1@uwaterloo.ca\\
\And
Chi-en Amy Tai\\
Systems Design Engineering\\ 
University of Waterloo\\
Waterloo, ON\\
amy.tai@uwaterloo.ca\\
\And
Alexander Wong\\
Systems Design Engineering\\ 
University of Waterloo\\
Waterloo, ON\\
alexander.wong@uwaterloo.ca\\
}

\author{Olivia Markham$^{1*}$ \quad Yuhao Chen$^{1*}$ \quad Chi-en Amy Tai$^{1}$ \quad Alexander Wong$^{1}$ \\
$^1$University of Waterloo, Waterloo, Ontario, Canada\\
$^*$Equal Contribution\\
}

\begin{document}

\maketitle

\begin{figure}[ht]
    \captionsetup[subfigure]{labelformat=empty, position=top, font=small, labelfont=bf}
    \centering
    \subfloat[LDM]{\frame{\includegraphics[width=0.24\linewidth]{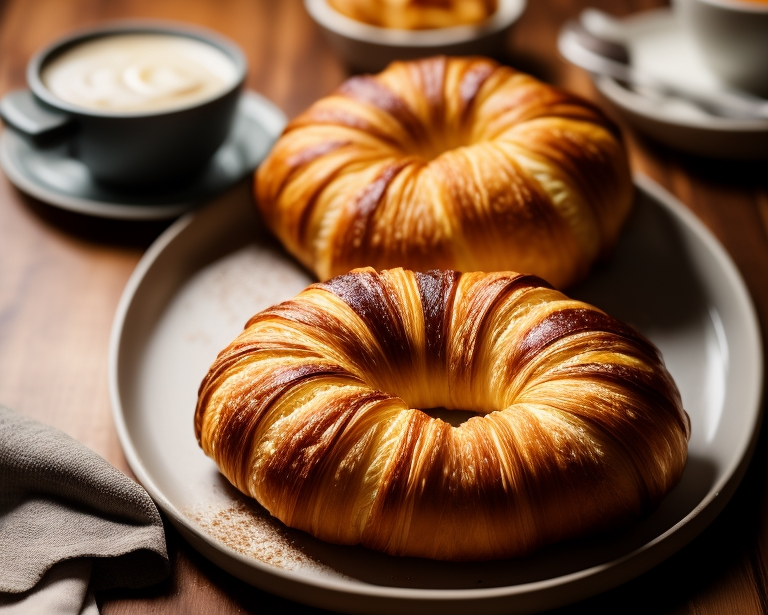}}}\hspace{2px}
    \subfloat[Midjourney]{\frame{\includegraphics[width=0.24\linewidth]{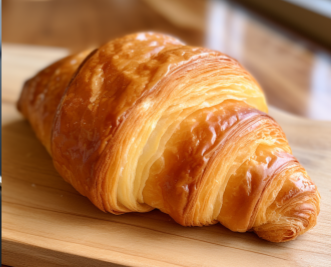}}}\hspace{2px}
    \subfloat[DALL-E Mini]{\frame{\includegraphics[width=0.24\linewidth]{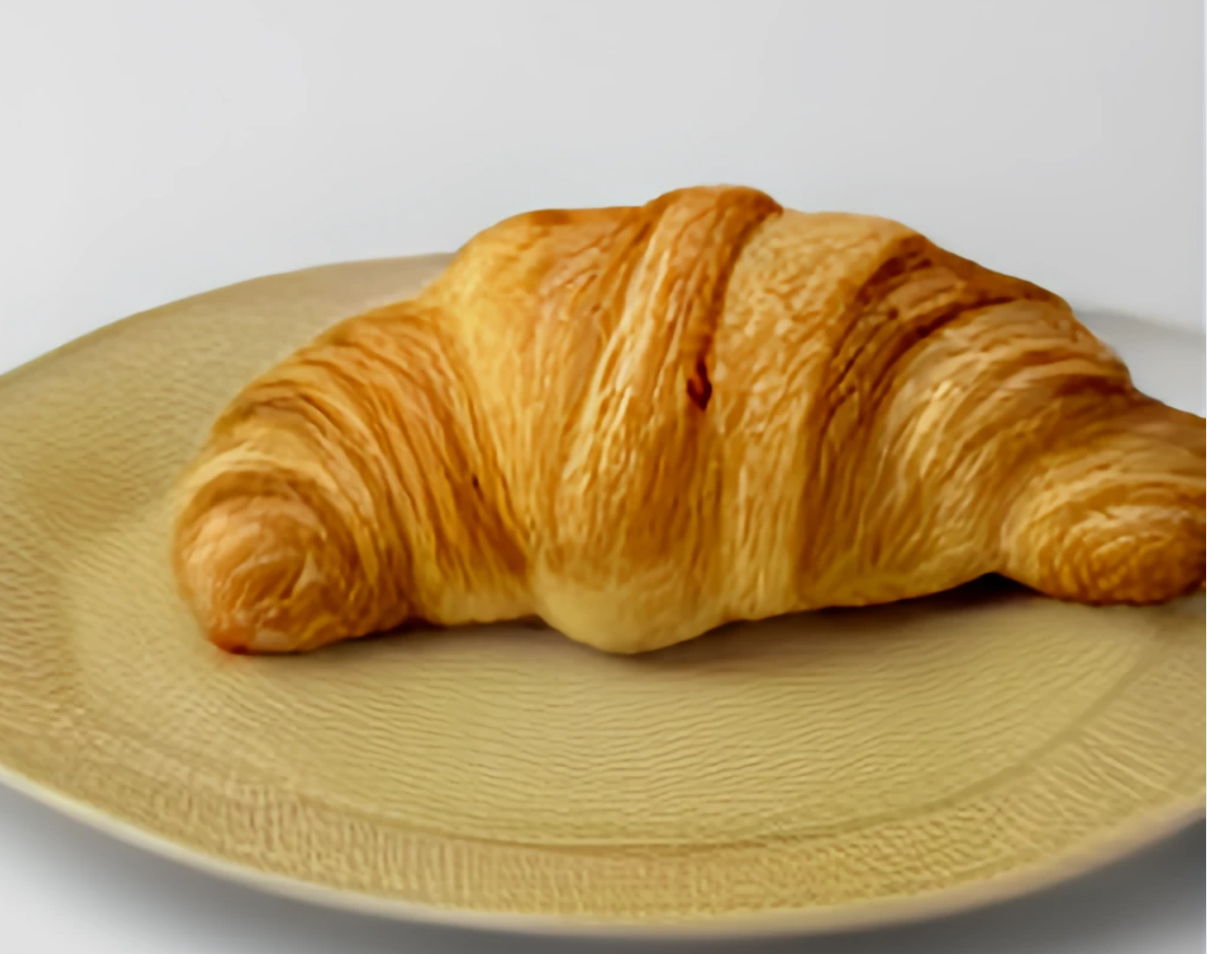}}}\hspace{2px}
    \subfloat[Ours]{\frame{\includegraphics[width=0.24\linewidth]{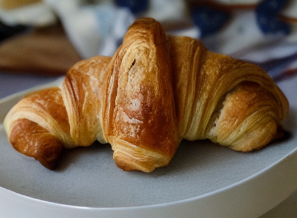}}}
    \caption{Sample image generation for the prompt "Croissant"}
    \label{fig:process}
\end{figure}

\begin{abstract}
Current state-of-the-art image generation models such as Latent Diffusion Models (LDMs) have demonstrated the capacity to produce visually striking food-related images. However, these generated images often exhibit an artistic or surreal quality that diverges from the authenticity of real-world food representations. This inadequacy renders them impractical for applications requiring realistic food imagery, such as training models for image-based dietary assessment. To address these limitations, we introduce FoodFusion, a Latent Diffusion model engineered specifically for the faithful synthesis of realistic food images from textual descriptions. The development of the FoodFusion model involves harnessing an extensive array of open-source food datasets, resulting in over 300,000 curated image-caption pairs. Additionally, we propose and employ two distinct data cleaning methodologies to ensure that the resulting image-text pairs maintain both realism and accuracy. The FoodFusion model, thus trained, demonstrates a remarkable ability to generate food images that exhibit a significant improvement in terms of both realism and diversity over the publicly available image generation models. We openly share the dataset and fine-tuned models to support advancements in this critical field of food image synthesis \href{https://bit.ly/genai4good}{here}.
\end{abstract}

\section{Introduction}

Generative image models have established themselves as effective tools for augmenting training data, leading to improved model performance and increased generalizability. This approach has particular relevance in generating life-like food images, which can greatly benefit applications like dietary assessment. However, our preliminary exploration of the widely recognized Latent Diffusion models \cite{rombach2021highresolution} revealed a challenge – these models did not consistently produce realistic food intake scenes. Upon closer examination, we identified the root cause: inconsistent label and image pairings within the Latent Diffusion Models (LDMs) training dataset\cite{rombach2021highresolution}.

In this paper, we address this limitation by training the Latent Diffusion model with better data. Our preliminary experiments demonstrated that when trained with high-quality, consistent data, Latent Diffusion could generate images virtually indistinguishable from ground truth. This revelation motivated us to develop FoodFusion, a latent diffusion model meticulously tailored for the authentic generation of food images based on textual descriptions. 

Our contributions encompass three key aspects: (1) \textbf{Comprehensive Dataset Compilation}: We have meticulously gathered an extensive array of datasets. These repositories encompass over 3,819 diverse food categories, comprising an impressive total of 335,386 images. (2) \textbf{Enhanced Data Quality}: We have pioneered the use of the Segment Anything Model (SAM)\cite{kirillov2023segany} to refine our image-caption pairs, enhancing their accuracy and ensuring a seamless integration between visual and textual data. (3) \textbf{Prompt Engineering to create LAIONFood dataset}: We have used prompt engineering to select realistic food images and created the LAIONFood dataset for training.

Through visual demonstration, we show that the FoodFusion model generates food images that stand out in both realism and consistency when compared to the LDMs.

\section{Optimal Data Yields Quality and Consistent Image Generation}

To evaluate the capabilities of the Latent Diffusion model,  we conducted an experiment where we trained the model on the NutritionVerse dataset \cite{tai2023nutritionverse}. This dataset consists of 361,997 meticulously curated images paired with consistent image-caption combinations, all exhibiting uniform lighting and scale within their respective scenes. The sole variations were in the food's composition and pose. Under these conditions, the trained Latent Diffusion model was adept at generating food images that closely resembled the original. This test highlights the model's potential to yield highly accurate images when provided with consistent and high-quality input data, thus inspiring us to develop an even more superior food dataset.

\section{Data Preparation for FoodFusion}

\begin{table}[ht]
  \caption{Food Dataset Information}
  \centering
  \begin{tabular}{lll}
    \toprule
    Dataset & Number of Images & Number of Categories \\
    \cmidrule(r){1-3}
    Food101\cite{bossard14} & 101,000 & 101 \\
    ISIA Food\cite{Min-ISIA-500-MM2020} & 130,469 & 251\\
    LAIONFood & 100,492 & 42 \\
    Food2k\cite{min2023large}& 1,036,722 & 2,000 \\
    Nutrition5k\cite{thames2021nutrition5k} & 3,425 & 3,425 \\
    \bottomrule
  \end{tabular}
  \label{food-dataset-info}
\end{table}

\textbf{Comprehensive Dataset Compilation}: To effectively train our FoodFusion model, we compiled a comprehensive list of open-sourced food datasets tailored for food recognition, summarized in Table \ref{food-dataset-info}. 
The Food2k dataset\cite{min2023large} contained watermarked image, causing the generation of images with distorted watermarks, so it was omitted from the training.
For Food101\cite{bossard14} and ISIA Food\cite{Min-ISIA-500-MM2020} datasets, we utilized the images and categories supplied as text prompts for corresponding images.
However, for the LAION-5B dataset (LAIONFood) and Nutrition5k\cite{thames2021nutrition5k}, the images and categories in these datasets were not immediately usable. Consequently, we present subsequent paragraphs on data enhancement and prompt engineering approaches, specifically designed to refine and optimize these datasets for our model's training.

\begin{figure}[ht]
    \captionsetup[subfigure]{labelformat=empty, position=top, font=small, labelfont=bf}
    \centering

    \newcommand{\subfigheight}{0.78in} 

    \subfloat[(A)]{\frame{\includegraphics[height=\subfigheight]{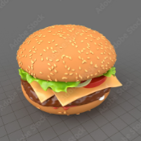}}}
    \hspace{4pt}
    \subfloat[(B)]{\frame{\includegraphics[height=\subfigheight]{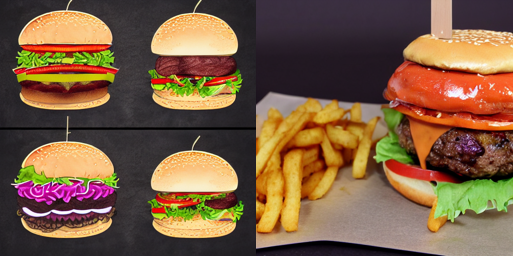}}}
    \hspace{7pt} 
    \subfloat[(C)]{\frame{\includegraphics[height=\subfigheight]{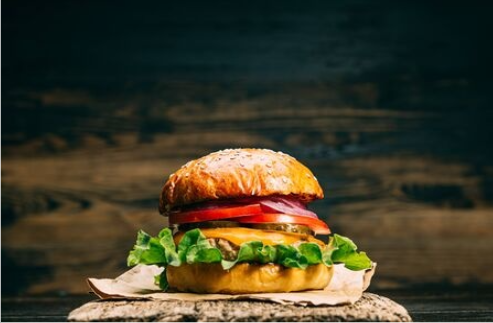}}}
    \hspace{4pt} 
    \subfloat[(D)]{\frame{\includegraphics[height=\subfigheight]{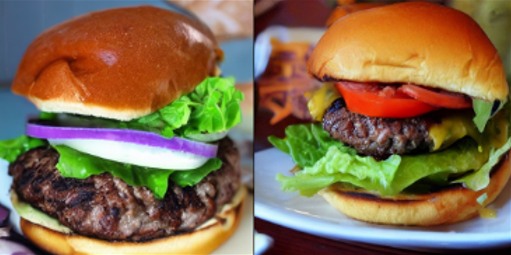}}}

    \caption{Querying and training Latent Diffusion models using Clip-retrieval of prompt of 'hamburger' in LAION5B vs. in LAIONFood. (A) the LAION5B training data. (B) the generated images from the LDM. (C) the LAIONFood training data  (D) the generated images from the FoodFusion model.}
    \label{fig:prompt_engineer}
\end{figure}

\textbf{Prompt Engineering for LAIONFood dataset}: Frequently, LDMs are trained on LAION5B, an extensive dataset consisting of image-caption pairs collected from the internet. However, a significant challenge associated with the LAION-5B dataset was the presence of non-realistic food images.For instance, searching for "apple" often yields drawings instead of photos. To tackle this issue, our research methodology integrated Clip-Retrieval with the LAION-5B dataset to create a diverse food image dataset. We initiated the process by querying LAION-5B with text prompts to obtain five high-quality images per food item, later expanding the dataset using image embeddings. This method produced 1,000 to 10,000 images for each of 42 food categories, encompassing major food groups inspired by \cite{mao2020visual}. The resulting LAIONFood dataset comprised 100,491 high-quality food images. Emphasizing data cleaning and prompt engineering, we used high aesthetic scoring and tailored image embedding prompts when querying with CLIP to enhance image quality and realism. Examples of cleaned data are shown in Figure \ref{fig:prompt_engineer}. 

\textbf{SAM-based Data Cleaning}: The Nutrition5k dataset\cite{thames2021nutrition5k} represents a collection of images paired with their corresponding ingredient lists. Initially, this dataset primarily contained raw food items along with their associated grammages and nutritional values, making it essential to undergo a comprehensive pre-processing phase. This involved removing imperceptible food items, such as salt and oil, and low-grammage ingredients for cleaner captions. Despite applying these thresholds, ongoing challenges prompted the implementation of GroundingDINO\cite{liu2023grounding} and SAM\cite{kirillov2023segany} to visually confirm ingredients, ensuring data consistency and accuracy.

\section{Experimental Results and Discussion}

We fine-tuned our model on the Latent Diffusion Model\cite{rombach2021highresolution} Our training process included a learning rate of $1.0 \times 10^{-0.5}$, utilizing a two A6000 GPU distributed setup for 28 epochs training on 335,386 data points. The resultant images, as depicted in Figure \ref{fig:process}, underscore the advanced ability of our FoodFusion model to generate more realistic food images in comparison to Latent Diffusion Model.

In this study, we have endeavored to contribute to the improvement of food image generation by introducing FoodFusion, addressing limitations seen in existing models like Latent Diffusion\cite{rombach2021highresolution}. Leveraging a vast dataset of over 300,000 curated image-caption pairs and fine-tuning specialized generative models, we have attained significant advancements in the consistent generation of realistic food images. Our future work will focus on training the Latent Diffusion model to produce food images with adjustable compositions and diverse viewing perspectives.

\begin{ack}
This work was supported by the National Research Council Canada (NRC) through the Aging in Place (AiP) Challenge Program, project number AiP-006.
\end{ack}

{
\small
\bibliography{new_in_ML}
}

\end{document}